\documentclass[conference]{IEEEtran}
\IEEEoverridecommandlockouts

\usepackage{titlesec}
\titleformat{\subsubsection}[runin] 
{\normalfont\itshape} 
 {\hspace{1em}\arabic{subsubsection})} 
{0.3em} 
{} 
\titlespacing*{\subsubsection}{0pt}{0.5em}{0.3em}  

\usepackage[dvipsnames]{xcolor}
\usepackage{soul}
\usepackage{threeparttable}
\usepackage[ruled,vlined,linesnumbered]{algorithm2e}
\usepackage{multirow}   
\usepackage{svg}
\usepackage{array}
\usepackage{booktabs}
\usepackage{siunitx}
\usepackage{tabularx}
\usepackage[inline,shortlabels]{enumitem} 

\usepackage{threeparttable}
\usepackage{float}
\usepackage{cite}
\usepackage{amsmath,amssymb,amsfonts}
\usepackage{graphicx}
\usepackage{textcomp}
\usepackage{ifthen}
\usepackage{url}

\graphicspath{{./figures/}{./figures/results}} 

\usepackage{cleveref} 
\Crefname{section}{Section}{Sections}
\crefname{figure}{Fig.}{Figs.}
\Crefname{figure}{Figure}{Figures}
\crefname{table}{Table}{Tables}
\Crefname{table}{Table}{Tables}

\usepackage[inline,shortlabels]{enumitem} 

\usepackage{makecell}

\def\eg{e.g., }

\def\trackchanges{1} 
\ifthenelse{\equal{\trackchanges}{1}}{
    
    \newcommand{\oldtext}[1]{\textcolor{purple}{\st{#1}} }
}{
    
    \newcommand{\oldtext}[1]{}
}

\begin{document}

\sisetup{detect-all} 

\title{Assessing the Generalization of Graph Neural Networks for Fault Location Across Increasing Distributed Energy Resource Penetration Levels}


\author{
\IEEEauthorblockN{
    Burak~Karabulut$^{\dagger,\star}$ \quad  
    Olayiwola~Arowolo$^\zeta$ \quad  
    Carlo~Manna$^{\star}$ \quad 
    Chris~Develder$^\dagger$ \quad 
    Jochen~L.~Cremer$^\zeta$
}
\vspace{5pt}

\vspace{5pt} 
\IEEEauthorblockA{%
\begin{tabular}{ccc}
$^\dagger$\textit{Dept.\ of Information Technology, IDLab}
  & $^\star$\textit{Water and Energy Transition Unit}
  & $^\zeta$\textit{Dept.\ Electrical Sustainable Energy} \\
\textit{Ghent University -- imec}, Ghent, Belgium
  & \textit{VITO}, Mol, Belgium
  & \textit{TU Delft}, Delft, Netherlands \\
\{burak.karabulut,\,chris.develder\}@ugent.be
  & \{burak.karabulut,\,carlo.manna\}@vito.be
  & \{o.a.arowolo,\,j.l.cremer\}@tudelft.nl
\end{tabular}}
}

\maketitle

\begin{abstract}
Accurate fault location is critical for distribution network reliability. However, increasing distributed energy resource (DER) penetration complicates fault location due to intermittent generation and bidirectional power flows that reshape fault signatures. Spatio-Temporal Graph Neural Networks (STGNNs) have shown promise by jointly modeling spatial and temporal dependencies, but their behavior under increasing DER penetration has not been studied rigorously. 
In this paper, we 
\begin{enumerate*}[(i)]
\item systematically benchmark spatio-temporal graph attention network (STGATv2) against purely temporal (gated recurrent unit, GRU), purely spatial (GATv2) and traditional machine learning baselines, and 
\item evaluate how well models generalize across increasing DER penetration levels (10\%, 25\%, 50\%) on a reconfigured IEEE 123-bus feeder with multiple DER injection points and moderate-to-high impedance faults.
\end{enumerate*}
Results show that STGATv2 consistently outperforms neural baselines, achieving 92--94\% macro F1 in-distribution. 
Notably, generalization across penetration levels is asymmetric: training at 50\% penetration retains near in-distribution F1 score at lower levels, whereas training at 10\% degrades considerably at 50\%\,---\,with STGATv2 retaining 81--84\% F1 under these drastic shifts, 
substantially higher than GATv2 and GRU which drop to 69--74\% F1 and 73--75\% F1 respectively.
Under realistic measurement noise, STGATv2 maintains $>85\%$ F1, while GRU drops as low as 33.5\% F1, highlighting the critical role of topological awareness for robust fault location in active distribution networks.

\end{abstract}


\section{Introduction}
\label{sec:intro}
Promptly locating faults within power distribution systems is essential for ensuring grid reliability and minimizing downtime~\cite{ieee1366}.
Identifying the faulty component\,---\,typically resulting from short circuits caused by environmental factors, hardware or insulation failure\,---\,enables operators to promptly isolate the affected area and restore service~\cite{grainger1994power}. 
 
Modern distribution networks are becoming larger and increasingly complex due to the increasing integration of distributed energy resources (DERs) and the electrification of demand, e.g., electric vehicle charging ~\cite{iea2023grids, liu2025graphgan}. DERs introduce inherent variability and intermittency, as weather-driven generation can lead to significant voltage fluctuations and load imbalances, while also altering fault propagation patterns. Specifically, bidirectional power flow enables fault currents to propagate from multiple directions rather than following a radial path within the grid~\cite{adetokun2023impact}. Consequently, fault location methods should remain accurate and robust with increasing penetration of DERs for reliable grid operation.

Existing fault location methods are generally categorized into model-based and data-driven approaches~\cite{rezapour2023review}.
Traditional \emph{model-based} techniques, such as impedance, voltage sag, and traveling wavelet methods~\cite{chandran2024extended, buzo2021new, liu2023novel}, rely on static assumptions regarding topology and fault characteristics. When operating conditions deviate from fixed parameters, these methods frequently suffer from increased modeling errors~\cite{mansourlakouraj2021application}. To overcome these limitations, \textit{data-driven methods} have been extensively studied~\cite{rezapour2023review}.  
Early machine learning (ML) approaches such as support vector machines (SVMs), and random forests (RF) use hand-crafted features to locate faults~\cite{chen2016fault}, which require  domain expertise and may not capture complex fault dynamics across changing grid conditions. More recent deep learning methods, such as Convolutional Neural Networks (CNNs) and Recurrent Neural Networks (RNNs), have shown potential by extracting features from measurement data, with CNNs capturing spatial correlations and RNNs modeling temporal dependencies~\cite{nguyen2023spatial}.
However, these methods often overlook the underlying grid topology and the non-uniform electrical connectivity between buses~\cite{chen2020fault}. This lack of topological awareness may limit performance, particularly in distribution networks with varying DER penetration levels.

To leverage the graph structure of distribution systems, Graph Neural Networks (GNNs) have emerged as a promising alternative for fault location~\cite{scarselli2009graph}.
By representing buses as graph nodes and grid lines as graph edges, GNNs are able to capture spatial dependencies across the feeder through the aggregation of information from neighboring buses, known as message passing~\cite{liao2022review}.
For example,~\cite{chen2020fault} applies Spectral Graph Convolutional Networks (GCNs), using the feeder’s Laplacian matrix in the graph Fourier domain to capture global structural information. 
Still, standard GNNs rely on fixed, structural normalization~\cite{kipf2016semi}, which can limit adaptability under changing network topologies.
Thus, Graph Attention Networks (GATs)~\cite{velickovic2018graph} and GraphSAGE~\cite{hamilton2017inductive} have been explored as robust alternatives for fault location~\cite{wang2024enhanced, Fan2024VGAEGSage} by weighting the importance of neighboring nodes or using localized aggregation through neighborhood sampling, respectively.

A limitation of standard GNN solutions for fault location is that they often ignore the inherent temporal dynamics of fault events. Consequently, Spatio-temporal GNNs (STGNNs)~\cite{nguyen2023spatial} have been proposed to jointly extract spatial and temporal dependencies, with recent extensions including multi-task graph-attention fault diagnosis~\cite{huang2027mtlgat} and related voltage-sag monitoring~\cite{pan2026gcngru} under specific DER penetrations. 
Despite the great potential shown by GNNs in fault location, the performance of state-of-the-art GNN models has not been explored in detail for power grids with varying DER penetration, which is increasing in present-day grids, and thus a highly relevant use case. Particularly, current literature has not covered
\begin{enumerate*}[(i)]
\item a rigorous performance comparison of the various models under different DER penetration levels since existing studies often consider few injection points, nor 
\item the generalization capability of GNN-based models to increasing DER penetration levels.
\end{enumerate*}
To address these gaps, this paper
\begin{enumerate}[(1)]
    \item Systematically and quantitatively benchmarks spatial-temporal GNN model against the spatial GNN, temporal GRU and traditional ML baselines under multiple DER injection points to assess the impact of joint spatio-temporal modeling; and
    \item Evaluates the generalization capabilities of the GNN-based approaches for fault location  with increasing and unseen DER penetration levels.
\end{enumerate}
The rest of the paper is organized as follows: 
\Cref{sec:method} presents the STGNN framework and its components, 
\Cref{sec:experiment-setup} discusses our experimental setup, and we present the results in 
\Cref{sec:results}
while 
\Cref{sec:conclusion} concludes the paper.

\section{Methodology}
\label{sec:method}
\begin{figure*}[t]
    \centering
\includegraphics[width=\textwidth]{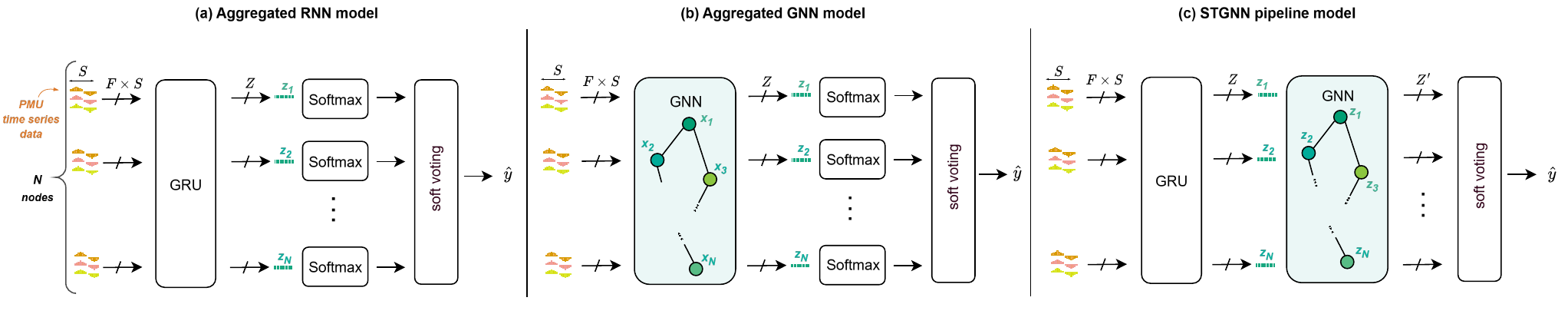}
    \caption{
    Model architectures for fault location: (a)~Shared GRU for temporal feature extraction per node; (b)~Shared GATv2, where measurement sequences are treated as features ($F_{in} = F \times S$) to capture spatial dependencies; and (c)~STGNN pipeline, where GRU-based temporal embeddings are refined via GNN  message passing. All models conclude with a classification head and soft voting to aggregate node-level probabilities. $N$: number of nodes, $F$: number of features, $S$: sequence length of measurement windows,
    $Z$: GRU hidden state dimension (and output for GRU(a) and GNN(b)), $Z'$: STGNN output dimension.}
    \label{fig:stgnn_architecture}
\end{figure*}

\subsection{Spatial Temporal Feature Extraction --- STGNN framework}
\label{subsec:stgnn}
As outlined in \Cref{sec:intro}, STGNNs have been proposed to jointly model spatial and temporal dependencies for fault location in distribution systems~\cite{nguyen2023spatial}. Specifically, this work adopts STGNN framework tailored for distribution grid fault location under increasing DER penetration levels and challenging fault conditions. The resulting pipeline (\cref{fig:stgnn_architecture}c) produces node-level temporal  representations  $\mathbf{z}_i$ via a GRU, which are processed through an improved GAT layer to extract spatial features, yielding $\mathbf{z}'_i$. These embeddings are then passed to a dense classifier to produce node-level predictions. At inference time, node-level outputs are aggregated using soft voting~\cite{ho1994decision}, by summing the class probabilities\,---\,including a ‘no fault’ case\,---\, across nodes. The final prediction $\hat{y}$ is obtained by selecting the class with the highest aggregate probability, effectively reducing the influence of outliers.


The graph \( G = (V, E) \) models the distribution network, where \( V \) is the set of \( N \) nodes (buses), i.e., \( |V| = N \), and \( E \) is the set of edges (lines) interconnecting them. This graph structure is defined by the adjacency matrix \( A \in \{0, 1\} ^{N \times N} \), where $A_{u,v} = 1$ if buses $u$ and $v$ are directly connected, and $A_{u,v} = 0$ otherwise.
However, the GNN topology does not necessarily need to be a 1-on-1 mapping of the full feeder. Instead, the graph is constructed using only measurement locations as nodes V, following the \emph{measured only} graph strategy proposed in~\cite{karabulut2024robustness}, reflecting the partial observability inherent in practical distribution systems.

\subsection{Temporal Feature Extraction -- Recurrent Neural Networks}
\label{sec:TemporalFeatureExtraction}
Fault events in distribution networks exhibit temporal behavior and form time series data. To extract these temporal features, the STGNN pipeline uses GRUs over LSTMs due to their simpler structure and lower computational cost, while maintaining sufficient capability to model the short-duration temporal dependencies typical of fault events. Specifically, a fixed-length window of per-phase root mean square (RMS) voltage measurements is input to the GRU, which produces a latent representation $\mathbf{z}_i$ for each node. These representations are passed to the GNN to incorporate spatial dependencies.

\subsection{Spatial Feature Extraction --  Graph Neural Networks}
\label{subsec:sfe_gnn}
In the GNN module, each node $v \in V$ is associated with a feature vector $h_v \in \mathbb{R}^{d}$, forming the node feature matrix $H \in \mathbb{R}^{N \times d}$, where each row corresponds to a bus in the distribution network. Although GNNs can incorporate edge features (\ e.g., line impedance or distance), in this work, only node features are considered. The objective of graph-based learning is then to
learn a mapping $\hat{y} = f(G; \theta)$ from the graph to a fault
location prediction, where $\theta$ denotes the learnable parameters.

GNNs learn node representations by iteratively aggregating information from neighboring nodes $u \in \mathcal{N}(v)$  across layers:
\begin{equation}
    H^{(k+1)} = f(H^{(k)}, A; \theta), \label{eq:gnn-general}
\end{equation}
where $H^{(k)} \in \mathbb{R}^{N \times d_k}$ represents the node representation matrix at the output of layer $k$. This matrix is formed by $h^{(k)}_v$, with each row representing a node $v \in V$ with a $d_k$-dimensional feature vector at layer $k$. Note that the neighborhood $\mathcal{N}(v)$ includes the node itself via self-loops ($v \in \mathcal{N}(v)$), ensuring that each node preserves its own features during aggregation. 

GNN architectures vary in how message passing is performed in \cref{eq:gnn-general}, particularly in how information from neighboring nodes is aggregated. 
In this work, we use an improved Graph Attention Network (GATv2)~\cite{brody2022attentive} to adaptively weight neighbors, as the relative importance of nodes shifts dynamically with the spatial distribution of DERs. Unlike fixed-weight GNNs~\cite{kipf2016semi}, this approach captures the evolving fault signatures inherent in active grids.
The representation of a node $v$ at layer $k+1$ is computed as: 
\begin{equation}
    h_v^{(k+1)} \;=\; \phi\!\left(
        \sum_{u \in \mathcal{N}(v)}
        \alpha_{vu}\, W\, h_u^{(k)}
    \right),
    \label{eq:gatv2-update}
\end{equation}
Here, $W$ is a learnable projection matrix and the attention
coefficient $\alpha_{vu} \in \mathbb{R}$ captures the relative
importance of neighbor $u$ for updating node $v$, and is calculated as:
\begin{equation}
    \alpha_{vu} = \frac{\exp\left( a^T \cdot \text{LeakyReLU}(W_1 \> h_v^{(k)} + W_2 \> h_u^{(k)}) \right)}{\sum_{j \in \mathcal{N}(v)} \exp\left( a^T \cdot \text{LeakyReLU}(W_1 \> h_v^{(k)} + W_2 \> h_j^{(k)}) \right)}, \label{eq:GATv2}
\end{equation}
where $a$ is a learnable attention vector and $\text{LeakyReLU}$ denotes a nonlinear activation. 
 
\section{Experimental Setup}
\label{sec:experiment-setup}

\subsection{Simulation Setup and Data Collection}
\label{sec:simulation-setup}

Due to the scarcity of real-world fault data and the limited observability of distribution networks, we generate synthetic data\footnote{Generating synthetic data is standard practice in distribution grid fault studies as real-world datasets are scarce and typically restricted due to security, and proprietary concerns~\cite{chen2020fault, nguyen2023spatial}.} via dynamic time series simulations in OpenDSS~\cite{epri_opendss} using PyDSS ~\cite{pydss}.
We use the IEEE 123-bus feeder, a standard benchmark in fault diagnosis studies~\cite{chen2020fault, nguyen2023spatial}, which operates at a frequency of \SI{60}{\hertz} and a nominal voltage of \SI{4.16}{\kilo\volt}.
For this study, the feeder is reconfigured by opening the tie switch at (60, 160) and closing the one at (54, 94).
Rerouting power through lateral branches with loads connected yields more subtle voltage variations across the measured nodes, thereby providing fault signatures that are more challenging to detect.

The  IEEE 123-bus feeder has a total active load of \SI{3.49}{\mega\watt}. To evaluate models' generalization capability  under increasing penetration levels, DERs, including PV, wind, and battery energy storage systems (BESS), are integrated into the reconfigured feeder at nominal penetrations of 10\%, 25\%, and 50\% of total load, representing current to near-future high-penetration deployment regimes\cite{eea2025renewables}. Higher generation is achieved by
progressively expanding the hosting node set, such that lower level configurations are subsets of higher ones. Consistent with typical deployment practices, DERs are first placed at single-phase lateral leaf nodes as small single-phase PV units and gradually extended toward central and upstream nodes with higher-capacity 3-phase units (3-phase PV, BESS, and wind) as aggregate generation increases. To assess the impact of spatial density, two systematic DER placement configurations are considered:
\begin{enumerate*}[(i)]
    \item \emph{Localized:} 
    a configuration with more spatially biased DER placement, mostly focusing on specific feeder sections at each penetration level and,
    \item \emph{Dispersed:} a configuration with a more uniform DER distribution across the network at each penetration level. Notably, in the local configuration, DER injections are primarily observed by a few proximate measurement nodes in the graph, whereas in the dispersed configuration, their impact is distributed across multiple monitoring points. (see \cref{table:dataset_info} for DER locations).\footnote{In the \emph{Localized} configuration, initially a larger set of hosting nodes (9 vs. 6) is used to increase spatial density while maintaining same aggregate penetration as the \emph{Dispersed} case. These partially distinct node sets provide varied injection patterns and prevent location bias.}
\end{enumerate*} 
    

All 11~short-circuit fault types~\cite{grainger1994power} are simulated across 25~locations. Fault duration is set to \SI{20}{\milli\second}, corresponding to the lower bound of typical primary protection clearing times. System responses are recorded at $N=25$ idealized measurement nodes with \SI{1}{\milli\second} resolution. We capture  three-phase RMS voltage magnitudes, as they are more accessible via existing infrastructure compared to current and
provide stable fault signatures across the feeder~\cite{nguyen2023spatial}. 
This results in sequences of 20~samples per fault event.

For each of the 25~fault scenarios, 100~simulations are run to capture a wide range of operating conditions.
Specifically, bus loads are independently scaled by $L \sim \mathcal{U}(\text{0.5, 1.3})$, representing off-peak to peak demand. Similarly, DER outputs are scaled using a shared factor $M \in \{\text{0, 0.25, 0.5, 0.75, 1}\}$, where $M$~=~1 corresponds to peak generation conditions, such as sunny weather with strong wind. BESS follow a probabilistic operating behavior conditioned on $M$. For $M$~$\geq$~0.5, the probabilities of charging, discharging, and idle operation respectively are 0.6\,/\,0.3\,/\,0.1, while for $M$~$<$~0.5, they are 0.3\,/\,0.6\,/\,0.1. This models higher charging activity under high generation and increased discharging under low generation. Fault resistances are sampled from $R_f \in \{\text{10, 20, 40, 80, 100}\}~\Omega$. Coupled with the switch reconfiguration, these moderate-to-high impedance faults yield subtle voltage drops (\num{1}--\SI{30}{\volt}), making fault location significantly more challenging than standard bolted faults.

From the resulting time series, \SI{59}{\milli\second} windows are extracted, producing 40 sliding windows of $S=20$ timesteps (\SI{20}{\milli\second}). The dataset contains \num{2.5} million samples (50\% no fault, 2\% per fault location), split at the simulation level into 70\% training, 15\% validation, and 15\% test sets, while preserving temporal alignment across measurement nodes.

\begin{figure}[t]
\centering
\includegraphics[width=1.04\columnwidth]{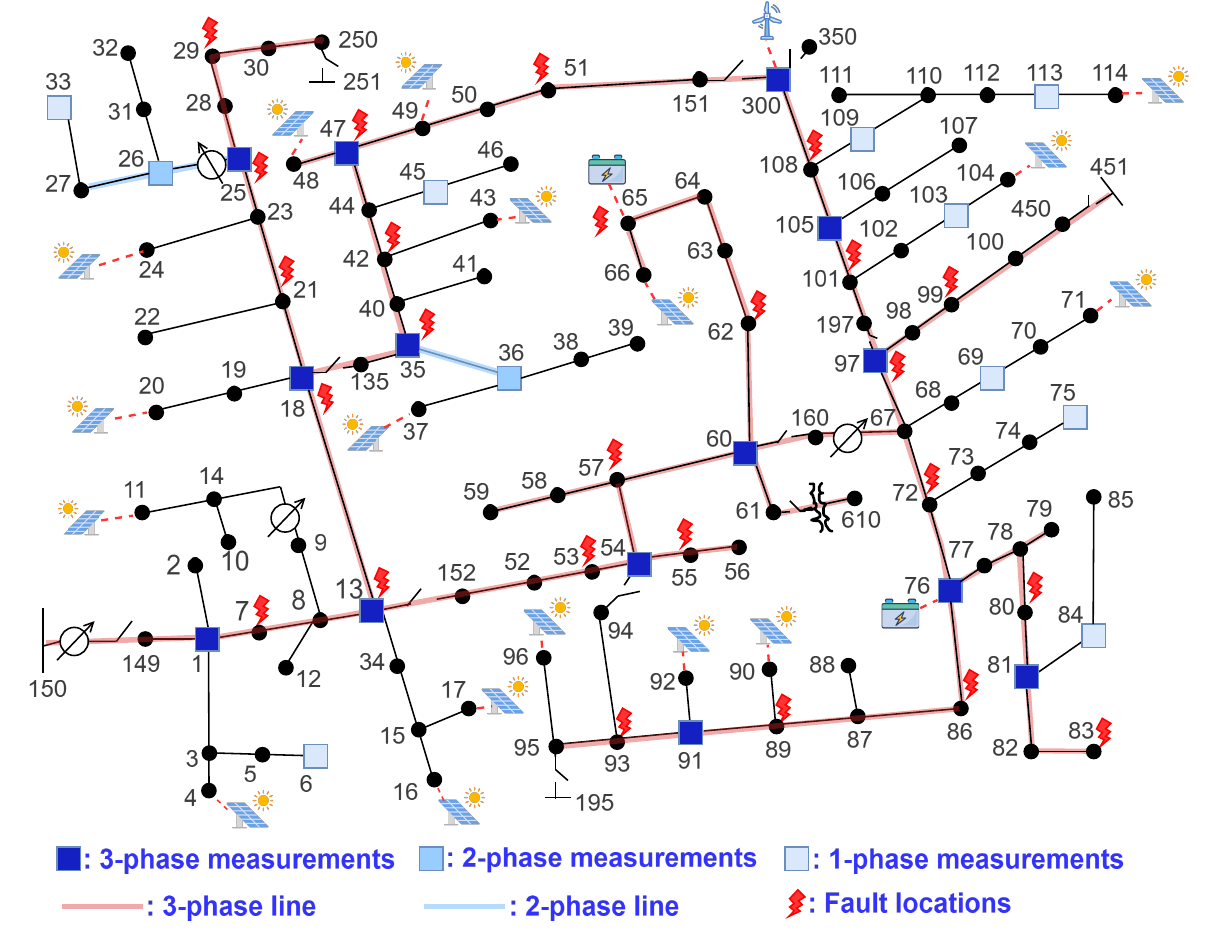} 
\caption{IEEE 123-node feeder with fault, DER, and measurement locations, shown for the 50\% penetration under dispersed configuration.}

\label{fig:IEEE123}
\end{figure}

\subsection{Model Training and Evaluation}
\label{sec:model-training}

Fault location is formulated as a \num{26}-class problem (\num{25} locations and a no-fault class) and optimized using cross-entropy loss. To evaluate the contribution of spatial and temporal modeling, we consider various baselines: \begin{enumerate*}[(i)]
\item traditional ML models, namely RF and SVM with Principal Component Analysis (PCA), to establish a performance reference on clean data;
    \item a purely \emph{temporal} model based on a GRU, where each measurement node is processed independently,
    \item a purely \emph{spatial} model based on GATv2 with residual connections, where each time window is flattened into a node feature vector ($F_{in} = F \times S$, see \cref{fig:stgnn_architecture}). 
\end{enumerate*}
All neural network models, including the STGNN, produce node-level predictions that are aggregated into graph-level outputs using a soft voting scheme, as described in \Cref{subsec:stgnn}.
Hyperparameters are selected empirically based on architectural characteristics. The STGNN and GRU baseline use a recurrent hidden size of \num{128} for temporal feature extraction. In the STGNN, this is followed by a GNN layer with a hidden dimension of \num{64}. The GATv2 baseline has a hidden dimension of \num{128}, which is then projected to \num{64} dimensions.

Neural network models are implemented in PyTorch, using PyTorch Geometric~\cite{fey2019fast} for GNN architectures, and trained on an Intel i5-9500 CPU with 64\ GB RAM. Input features $(V_1, V_2, V_3)$ are Z-score normalized. Node embeddings are mapped to \num{26}-dimensional logits via a fully connected layer, with ReLU activations. During training, PyTorch’s cross-entropy loss internally applies log-softmax for numerical stability. We apply dropout of \num{0.35} and batch normalization; GATv2 models use attention dropout of \num{0.3} with four heads. All models are trained until convergence, using AdamW 
with learning rate \num{0.0005} and weight decay \num{1e-4}. Performance is evaluated using the macro F1-score on the test set.

\begin{table}[t]
\caption{Bus indices for fault, measurement device and DER placement. DERs added to the subsets are typeset in \textcolor{green!75!black}{green}.}
\centering
\resizebox{\columnwidth}{!}{
\begin{tabular}{lccc}
\toprule
\toprule
\textbf{Elements} & \multicolumn{2}{l}{\textbf{Bus Locations}} & \textbf{Number} \\
\midrule
Fault Nodes & \multicolumn{2}{p{6cm}}{7, 13, 18, 21, 25, 29, 35, 42, 47, 51, 53, 55, 57, 62, 65, 72, 80, 83, 86, 89, 93, 97, 99, 101, 108} & 25 \\
\midrule
Measured Nodes & \multicolumn{2}{p{6cm}}{1, 6, 13, 18, 25, 26, 33, 35, 36, 45, 47, 54, 60, 69, 75, 76, 81, 84, 91, 97, 103, 105, 109, 113, 300} & 25 \\
\bottomrule
\toprule
\multicolumn{3}{c}{\textbf{Localized DER Configuration}} \\
\midrule
10\% Penetration & \multicolumn{2}{p{6cm}}{4, 11, 20, 24, 37, 43, 71, 88, 104} & 9 \\
\cmidrule{1-4}
25\% Penetration & \multicolumn{2}{p{6cm}}{4, 11, \textcolor{green!75!black}{16}, \textcolor{green!75!black}{17}, 20, 24, 37, 43, \textcolor{green!75!black}{48}, 71, \textcolor{green!75!black}{76}, 88, 104} & 13 \\
\cmidrule{1-4}
50\% Penetration & \multicolumn{2}{p{6cm}}{4, 11, 16, 17, 20, 24, 37, 43, 48, \textcolor{green!75!black}{49}, \textcolor{green!75!black}{65}, \textcolor{green!75!black}{66}, 71, 76, 88, \textcolor{green!75!black}{90}, \textcolor{green!75!black}{92}, \textcolor{green!75!black}{96}, 104, \textcolor{green!75!black}{300}} & 20 \\
\midrule
\midrule
\multicolumn{3}{c}{\textbf{Dispersed DER Configuration}} \\
\midrule
10\% Penetration & \multicolumn{2}{p{6cm}}{12, 24, 43, 66, 71, 88} & 6 \\
\cmidrule{1-4}
25\% Penetration & \multicolumn{2}{p{6cm}}{12, \textcolor{green!75!black}{16}, 24, \textcolor{green!75!black}{39}, 43, \textcolor{green!75!black}{48}, 66, 71, \textcolor{green!75!black}{76}, 88, \textcolor{green!75!black}{114}} & 11 \\
\cmidrule{1-4}
50\% Penetration & \multicolumn{2}{p{5cm}}{\textcolor{green!75!black}{11}, 12, 16, 24, \textcolor{green!75!black}{32}, 39, 43, 48, \textcolor{green!75!black}{59}, \textcolor{green!75!black}{65}, 66, 71, 76, \textcolor{green!75!black}{85}, 88, \textcolor{green!75!black}{92}, \textcolor{green!75!black}{96}, 114, \textcolor{green!75!black}{300}} & 19 \\
\bottomrule
\end{tabular}
}
\label{table:dataset_info}
\end{table}

\section{Results and Discussion}
\label{sec:results}
\begin{table*}[t]
\centering
\caption{Fault Location Macro F1 (\%) $\pm$ Standard Deviation Across 3 random seeds, by Train/Test DER Penetration Levels}
\label{table:unified_wide}
\renewcommand{\arraystretch}{1.2}
\setlength{\tabcolsep}{5pt} 

\newcolumntype{C}{>{\centering\arraybackslash}p{1.7cm}}
\begin{tabular}{l c CCC c CCC}
\toprule
 & & \multicolumn{3}{c}{\textbf{Localized DER Configuration}} & & \multicolumn{3}{c}{\textbf{Dispersed DER Configuration}} \\
 \cmidrule(lr){3-5} \cmidrule(lr){7-9}
\textbf{Model} & \textbf{\makecell{Training (\%)}} & \multicolumn{3}{c}{\textbf{Test (\%)}} & & \multicolumn{3}{c}{\textbf{Test (\%)}} \\
\cmidrule(lr){3-5} \cmidrule(lr){7-9}
 & & \textbf{10} & \textbf{25} & \textbf{50} & & \textbf{10} & \textbf{25} & \textbf{50} \\
\midrule
\multirow{3}{*}{STGATv2} & 10 & \textbf{94.07 {\scriptsize $\pm$ 1.82}} & 92.97 {\scriptsize $\pm$ 1.53} & 81.47 {\scriptsize $\pm$ 0.97} & & \textbf{94.03 {\scriptsize $\pm$ 0.24}} & 93.93 {\scriptsize $\pm$ 0.20} & 83.53 {\scriptsize $\pm$ 0.89} \\
                         & 25 & 93.77 {\scriptsize $\pm$ 0.37} & \textbf{93.63 {\scriptsize $\pm$ 0.42}} & 85.77 {\scriptsize $\pm$ 0.72} & & 93.73 {\scriptsize $\pm$ 0.46} & \textbf{94.03 {\scriptsize $\pm$ 0.10}} & 85.90 {\scriptsize $\pm$ 2.05} \\
                         & 50 & 90.70 {\scriptsize $\pm$ 0.71} & 89.60 {\scriptsize $\pm$ 0.85} & \textbf{92.37 {\scriptsize $\pm$ 0.55}} & & 91.67 {\scriptsize $\pm$ 2.44} & 91.03 {\scriptsize $\pm$ 2.90} & \textbf{93.03 {\scriptsize $\pm$ 1.07}} \\
\midrule
\multirow{3}{*}{GATv2}   & 10 & \textbf{85.83 {\scriptsize $\pm$ 2.05}} & 84.53 {\scriptsize $\pm$ 2.40} & 73.93 {\scriptsize $\pm$ 2.41} & & \textbf{85.50 {\scriptsize $\pm$ 1.66}} & 83.93 {\scriptsize $\pm$ 1.81} & 69.30 {\scriptsize $\pm$ 2.19} \\
                         & 25 & 86.70 {\scriptsize $\pm$ 1.95} & \textbf{86.13 {\scriptsize $\pm$ 1.50}} & 75.37 {\scriptsize $\pm$ 2.13} & & 84.17 {\scriptsize $\pm$ 0.97} & \textbf{83.83 {\scriptsize $\pm$ 0.77}} & 77.63 {\scriptsize $\pm$ 3.21} \\
                         & 50 & 83.30 {\scriptsize $\pm$ 3.05} & 82.73 {\scriptsize $\pm$ 3.38} & \textbf{86.70 {\scriptsize $\pm$ 2.70}} & & 86.23 {\scriptsize $\pm$ 2.40} & 85.33 {\scriptsize $\pm$ 2.18} & \textbf{87.93 {\scriptsize $\pm$ 1.90}} \\
\midrule
\multirow{3}{*}{GRU}     & 10 & \textbf{84.53 {\scriptsize $\pm$ 2.85}} & 83.73 {\scriptsize $\pm$ 2.94} & 73.50 {\scriptsize $\pm$ 2.36} & & \textbf{82.47 {\scriptsize $\pm$ 2.47}} & 81.40 {\scriptsize $\pm$ 1.91} & 74.73 {\scriptsize $\pm$ 2.06} \\
                         & 25 & 77.57 {\scriptsize $\pm$ 3.46} & \textbf{74.43 {\scriptsize $\pm$ 4.16}} & 70.70 {\scriptsize $\pm$ 0.57} & & 83.07 {\scriptsize $\pm$ 2.82} & \textbf{82.73 {\scriptsize $\pm$ 2.98}} & 76.10 {\scriptsize $\pm$ 3.06} \\
                         & 50 & 76.83 {\scriptsize $\pm$ 6.31} & 76.53 {\scriptsize $\pm$ 6.76} & \textbf{76.53 {\scriptsize $\pm$ 6.30}} & & 85.27 {\scriptsize $\pm$ 1.59} & 84.03 {\scriptsize $\pm$ 1.80} & \textbf{85.13 {\scriptsize $\pm$ 1.44}} \\
\bottomrule
\end{tabular}
\end{table*}

\subsection{Comparing Spatio-Temporal GNN with baseline models}
\label{subsec:performance_comparsion}

\cref{table:unified_wide} reports fault location performance for all neural models across DER penetration levels for both configurations. While traditional ML baselines achieve high in-distribution performance (87--93\% F1 for RF; 83--87\% F1 for PCA-SVM), their inherent sensitivity to distribution shifts leads to severe degradation under unseen DER penetration levels\footnote{Under increasing DER levels (\eg train on 10\% and test on 50\%), F1 scores drop to as low as 47\% (RF) and 60\% (PCA-SVM).}, precluding them from detailed discussion in subsequent subsections.
Across both configurations, STGATv2 maintains in-distribution F1 scores~(diagonals in \cref{table:unified_wide}) roughly 5-10 points higher than GATv2 and 8-19 points higher than GRU baselines.
This performance gap suggests that jointly modeling spatial and temporal dependencies is more informative than either alone: while the GRU captures the temporal dynamics of fault-induced voltage drop, it lacks the topological awareness to correlate these drops across the feeder topology. Conversely, GATv2 leverages graph topology to capture fault signatures but lacks the temporal memory to distinguish them from load variations. By extracting temporal features before propagating them via attention-based aggregation, STGATv2 jointly captures spatio-temporal characteristics of active networks. Looking at in-distribution performance more closely, GATv2 consistently outperforms GRU across both configurations (e.g., 86.13\% vs.\ 
74.43\% F1 at 25\% \emph{localized}, with the gap most pronounced under higher penetration in this setting), suggesting that attention-based message passing over flattened temporal features may provide a stronger inductive bias than sequential temporal processing alone.

In particular, STGATv2 and GRU achieve a higher F1 score in the \emph{dispersed} configuration, particularly at higher training penetration levels, as its uniform DER distribution ensures that fault-induced voltage drops are observed more consistently across measurement nodes. Furthermore, having fewer DER hosting nodes requires higher per-node injections that act as significant negative loads. This leads to more pronounced voltage shifts during faults, making signatures easier to capture. However, GATv2 shows no consistent direction between configurations, with slightly better performance in the 25\% \emph{localized} in-distribution case suggesting that spatial locality can provide a concentrated feature for attention mechanisms to leverage, while the uniform distribution of the \emph{dispersed} configuration can dilute this spatial focus.

\subsection{Asymmetric Generalization to DER Penetration Level}
Beyond superior in-distribution performance, STGATv2 generalizes better across varying penetration levels compared to baseline models.
However, this generalization is inherently asymmetric: models trained at 50\% penetration retain their performance at lower levels, dropping at most 4 points. 
Conversely, training at 10\% for 50\% testing leads to substantial degradation; while STGATv2 in the \emph{localized} configuration drops to 81.47\% F1, GATv2 and GRU drop more significantly to 73.93\% and 73.50\% F1, respectively. 
Interestingly, this drop is not gradual, as models trained at 10\% lose at most 1.6 points at 25\% whereas 16.2 at 50\%, for our configurations.
This suggests high penetration, with complex bidirectional flows, includes fault patterns seen at lower levels. Thus, training at high penetration provides a diverse feature set that generalizes downward, while models trained at low penetration fail to adapt to complex signatures under severe DER shifts.

Notably, the models generalize more effectively in the \emph{dispersed} configuration compared to the \emph{localized} setup. Consistent with the discussion in \Cref{subsec:performance_comparsion}, the uniform DER distribution provides relatively more stable measurement patterns across measurement nodes. This allows	STGATv2	to retain 83.53\% F1 in	the	dispersed configuration	versus	81.47\%	in localized under	the	10\% to 50\% shift, whereas GRU experiences a narrower gap between	configurations (74.73\% F1 vs. 73.50\% F1).
In contrast, GATv2 shows a larger gap between configurations under this severe shift, landing at 69.30\% F1 in the dispersed configuration versus 73.93\% in localized, as its attention mechanism may become diluted across distributed DER signatures. Ultimately, integrating spatial reasoning with temporal dynamics is essential to maintain fault location performance as DER penetration increases.

\subsection{Impact of Measurement Noise on Model Performance}
Following the assessment of generalization capabilities, we investigate model robustness to measurement noise while maintaining baseline DER penetration levels, isolating noise as the sole source of distribution shift. This is particularly critical as moderate to high fault impedance combined with the reconfigured feeder leads to weaker fault signatures, yielding notably subtle voltage drops (\num{1}--\SI{30}{\volt}). Measurement noise is introduced via zero-mean Gaussian perturbations 
with signal-to-noise ratios (SNRs) of \SI{50}{\decibel} and \SI{45}{\decibel}. The standard deviation of the noise is calculated as $\sigma_\text{noise} = \text{10}^{-\text{SNR}/\text{20}}$. Noise is injected for each sample in the batch independently, resulting in unseen test perturbations.

At \SI{50}{\decibel}, model performance decreases  across the board. STGATv2 remains the most consistent with an F1 drop of up to $6$ points, maintaining performance between 88\% and 90\% F1 across both configurations. The purely spatial GATv2 model drops up to $7$ points, landing at around 81\% F1. On the other hand, the sequential GRU shows higher sensitivity to noise; its performance falls up to 15 points to 70\%--72\% F1 on \emph{localized} and up to 23 points to 62.5\% F1 on \emph{dispersed}.
This sharper decline occurs because GRU lacks spatial awareness to isolate subtle fault-induced voltage drops masked by noise.
When noise is increased to \SI{45}{\decibel}, the importance of topological awareness becomes more evident. STGATv2 retains a score between 85\% and 87\% F1, as the joint spatio-temporal modeling allows it to effectively filter out the measurement noise. GATv2 experiences a higher drop compared to STGATv2 as it lacks the temporal memory needed to distinguish the fault dynamics, landing at around $74\%$ F1. Meanwhile, 
GRU performance drops significantly, falling to 62\% F1 (\emph{localized}) and 33.5\% F1 (\emph{dispersed}). Surprisingly, GRU suffers a steeper F1 drop in the \emph{dispersed} configuration under measurement noise. This may be attributed to uniform DER distribution, combined with the noise masking already subtle fault signatures across the network, whereas, in the localized configuration, nodes away from the DER injection points remain more distinguishable.





In summary, we observe that topological (spatial) awareness appears to be fundamental for model robustness in distribution grid fault location, especially under subtle fault signal conditions. While modeling both spatial and temporal correlations provides the highest robustness to noise, the temporal GRU model struggles to identify fault patterns once the signal-to-noise ratio degrades local measurement integrity. 

\section{Conclusion and Future Work}
\label{sec:conclusion}
This work systematically 
\begin{enumerate*}[(i)]
\item benchmarks STGATv2 against purely spatial, purely temporal, and traditional ML baselines under multiple DER injection points to assess the impact of joint spatio-temporal modeling, and
\item  assesses model generalization to increasing and unseen DER penetration levels for distribution network fault location.
\end{enumerate*}
Our results confirm that jointly modeling spatial and temporal dependencies outperforms baselines (92--94\% F1). Notably, generalization across DER penetration is asymmetric: models trained at high penetration (50\%) retain near in-distribution performance at lower levels, whereas models trained at low penetration (10\%) degrade severely under high penetration shifts. Despite this degradation, STGATv2 retains 81–84\% F1, while baselines drop sharply.
Furthermore, under realistic measurement noise, STGATv2 maintains robust performance ($>85\%$ F1) compared to severe baseline degradation, suggesting that topological awareness is critical for robustness. Future work includes extending this framework to larger, diverse networks to evaluate model generalization and sensitivity to DER placement configurations. Additionally, evaluating extreme DER penetration regimes (up to 100\%) will provide insights into robustness under severe voltage volatility and operational limits, while incorporating detailed grid and inverter-based dynamics will support validation toward real-world deployment.

\section*{Acknowledgment}
Research reported in this publication was supported by VITO grant number VITO\_UGENT\_PhD\_2301 and partially funded by the Flemish Government (under the 'Onderzoeksprogramma Artificiële Intelligentie (AI) Vlaanderen' programme). This research was also supported by the Research Foundation - Flanders (FWO) under grant number V425326N.

\bibliographystyle{ieeetr} 
\bibliography{references}

\newpage 
\onecolumn

\end{document}